\documentclass[11pt]{article}
\usepackage{multirow}
\usepackage[final]{acl}
\usepackage{times}
\usepackage{latexsym}

\usepackage[T1]{fontenc}

\usepackage[utf8]{inputenc}

\usepackage{microtype}
\usepackage{booktabs}
\usepackage{url}
\usepackage{hyperref}
\usepackage{inconsolata}
\usepackage{caption}

\usepackage{graphicx}
\usepackage{comment}
\usepackage{fvextra}

\usepackage{tikz}
\usetikzlibrary{shapes.geometric, arrows.meta, positioning, calc}
\usepackage{pifont}
\usepackage{xcolor}
\usepackage{soul}
\usepackage{graphicx}

\definecolor{quoteBg}{RGB}{237,242,250}
\definecolor{quoteBorder}{RGB}{190,205,225}

\definecolor{retrGreen}{RGB}{15,110,86}
\definecolor{wikiBg}{RGB}{225,245,238}
\definecolor{wikiBorder}{RGB}{93,202,165}
\definecolor{rejBg}{RGB}{243,243,240}
\definecolor{rejBorder}{RGB}{190,190,184}
\definecolor{rejText}{RGB}{140,140,134}

\definecolor{compBlue}{RGB}{24,95,165}
\definecolor{hlBlueBg}{RGB}{219,234,250}

\definecolor{llmMaroon}{RGB}{153,53,86}
\definecolor{baseBg}{RGB}{250,236,231}
\definecolor{baseBorder}{RGB}{240,153,123}
\definecolor{baseText}{RGB}{153,60,29}
\definecolor{ragBg}{RGB}{234,243,222}
\definecolor{ragBorder}{RGB}{151,196,89}
\definecolor{ragText}{RGB}{59,109,17}

\sethlcolor{hlBlueBg}

\DefineVerbatimEnvironment{prompt}{Verbatim}{
  fontsize=\tiny,
  breaklines=true,
  breakanywhere=true,
  breaksymbolleft={},
  breaksymbolright={},
  commandchars=\\\{\}
}

\title{RAGing Against Misogyny: Improving Implicit Misogyny Explanations}

\author{
  \textbf{Ivana Crescenzi\textsuperscript{1,\textdagger}},
  \textbf{Arianna Muti\textsuperscript{1,\textdagger}},
  \textbf{Maarten Sap\textsuperscript{2}},
  \textbf{Tommaso Caselli\textsuperscript{3}},
  \textbf{Debora Nozza\textsuperscript{1}}
  \\
  \textsuperscript{1}Department of Computing Sciences, Bocconi University, Milan, Italy
  \\
  \textsuperscript{2}Language Technologies Institute, Carnegie Mellon University, Pittsburgh, USA
  \\
  \textsuperscript{3}Center for Language and Cognition Groningen, University of Groningen, The Netherlands
  \\
  {\small\textsuperscript{\textdagger}These authors contributed equally.}
}

\begin{document}
\maketitle
\begin{abstract}
Implicit misogyny often requires background knowledge to be understood.
To address this challenge, we propose a Retrieval-Augmented Generation (RAG) framework for detecting and explaining implicit misogyny in English and Italian, leveraging encyclopedic, slang and feminist knowledge sources. 
Using the ImplicIT-Mis and SBIC+ datasets, we evaluate the contribution of each knowledge source to detection and explanation quality, investigating also whether RAG addresses known LLM failure modes in implicit misogyny.
Results show that 
RAG improves recall on ImplicIT-Mis, but does not improve detection performance on \textsc{SBIC+}. However, human evaluation shows that RAG-based explanations are preferred over no-RAG explanations in both languages. Feminist literature emerges as the most consistently useful source for detection and explanation.


 \noindent {\color{red}{\textbf{Content warning}}}: This paper contains examples of misogynistic language.
\end{abstract}

\section{Introduction}
\label{sec:introduction}
Online misogyny constitutes a digital manifestation of patriarchal power, functioning to undermine women’s credibility and limit their participation in the online sphere \cite{jurasz2019online}. 
Misogynistic language is expressed not only through explicit insults, but also through indirect, coded, and context-dependent language that normalizes hostility toward women while often remaining deniable \cite{jankowicz2021malign}. 

Recent work has increasingly examined the implicit realization of hate speech and abusive language \cite{caselli-etal-2020-feel,wiegand-etal-2021-implicitly-abusive,ocampo-etal-2023-depth,muti-etal-2024-language}.  
Understanding implicit misogyny often requires background knowledge to interpret sarcasm, cultural references, pragmatic cues, and insinuations about women’s competence, sexuality, emotionality, or place in public life. Without it,
LLMs tend to rely on shallow semantic associations, producing generic or unsupported explanations \cite{muti-etal-2024-language, muti-etal-2025-r}.
Such limitations matter for moderation: while structured explanations can make moderation decisions faster and more usable for moderators \citep{zhang-etal-2023-biasx, calabrese-etal-2024-explainability}, 
content moderation systems have been shown to under-moderate implicit hate speech \cite{hartmann2025lost}.

\begin{figure}[!t]
\centering
\resizebox{0.97\columnwidth}{!}{%
\begin{tikzpicture}[
  font=\fontsize{8}{9.5}\selectfont,
  box/.style={rectangle, rounded corners=2pt, draw=black!40, thin, align=left,
              inner sep=4pt, outer sep=0pt, anchor=north west},
  sec/.style={anchor=north west, outer sep=0pt, inner sep=0pt}
]

\node[box, fill=quoteBg, draw=quoteBorder, text width=7.6cm] (input) at (0,0) {
{\fontsize{7.2}{8.6}\selectfont\color{black!55}\textbf{Post:}\ }\textit{``Dignified is silence, not the spectacularization of an announced tragedy. You are killing the other murdered women again; remain in the dignified silence of relatives. But the show must go on, right?''}\hspace{4pt}
{\fontsize{7.2}{8.6}\selectfont\color{black!55}\textbf{Context:} \textit{``Woman of the year? One name could be \colorbox{wikiBg}{\textcolor{retrGreen}{\bfseries Elena Cecchettin}}, Giulia's sister: an example of dignity, a tragic voice trying to turn the negative into something positive.''} }
};

\node[sec, text=retrGreen] (lab1) at ($(input.south west)+(0,-0.8mm)$)
     {{\fontsize{10}{10}\selectfont\bfseries 1}\,\textemdash\,{\fontsize{8.5}{8.5}\selectfont\bfseries RAG RETRIEVAL}};

\node[box, fill=wikiBg, draw=wikiBorder, text width=4.78cm, minimum height=1.05cm] (wiki) at ($(lab1.south west)+(0,-0.7mm)$) {
\textbf{\color{retrGreen}\ding{51}\ Wikipedia}\\
{\fontsize{6.9}{8.2}\selectfont\ttfamily wiki:Murder\_of\_Giulia\_Cecchettin}
};
\node[box, fill=rejBg, draw=rejBorder, text=rejText, text width=0.9cm, minimum height=1.05cm] (slang) at ($(wiki.north east)+(1.5mm,0)$) {Slang};
\node[box, fill=rejBg, draw=rejBorder, text=rejText, text width=1.05cm, minimum height=1.05cm] (books) at ($(slang.north east)+(1.5mm,0)$) {Feminist books};

\node[sec, text=compBlue] (lab2) at ($(wiki.south west)+(0,-1.2mm)$)
     {{\fontsize{10}{10}\selectfont\bfseries 2}\,\textemdash\,{\fontsize{8.5}{8.5}\selectfont\bfseries RAG COMPRESSION}};

\node[box, dashed, draw=black!35, fill=black!2, text width=7.6cm] (fullwiki) at ($(lab2.south west)+(0,-0.7mm)$) {
{\fontsize{7.2}{8.6}\selectfont\ttfamily wiki:Murder\_of\_Giulia\_Cecchettin}\\[2pt]
{\fontsize{7.6}{9.2}\selectfont\color{black!60}
Giulia Cecchettin, an Italian student, was murdered by her ex-boyfriend on 11 November 2023, sparking nationwide protests against femicide. [\dots]~
\textcolor{compBlue}{\hl{\textbf{On 20 November 2023, the Corriere della Sera published a letter by Elena Cecchettin denouncing society's role in producing ``healthy sons of patriarchy and rape culture.''}}}
~[\dots]}
};

\node[sec, text=llmMaroon] (lab3) at ($(fullwiki.south west)+(0,-1.2mm)$)
     {{\fontsize{10}{10}\selectfont\bfseries 3}\,\textemdash\,{\fontsize{8.5}{8.5}\selectfont\bfseries LLM EXPLANATION}};

\node[box, fill=baseBg, draw=baseBorder, minimum width=3.86cm, minimum height=1.7cm] (base) at ($(lab3.south west)+(0,-0.7mm)$) {};
\node[align=left, text width=3.4cm, anchor=north west] at ([xshift=4pt,yshift=-4pt]base.north west) {
{\fontsize{7.2}{8.6}\selectfont\bfseries\color{baseText}\ding{55}\ BASELINE (NO RAG)}\\[2pt]
{\fontsize{7.6}{9.2}\selectfont Women are exploited to attract audiences and attention.}
};

\node[box, fill=ragBg, draw=ragBorder, minimum width=3.86cm, minimum height=1.7cm] (rag) at ($(base.north east)+(1.5mm,0)$) {};
\node[align=left, text width=3.6cm, anchor=north west] at ([xshift=4pt,yshift=-4pt]rag.north west) {
{\fontsize{7.2}{8.6}\selectfont\bfseries\color{ragText}\ding{51}\ RAG-GENERATED}\\[2pt]
{\fontsize{7.6}{9.2}\selectfont Women who speak publicly about their suffering are not dignified.}
};

\end{tikzpicture}%
}
\caption{Example RAG pipeline. Among the candidate sources, the relevant Wikipedia page is retrieved; compression retrieves only the key sentences within the full article, enabling the model to draw the correct implied assumption (unlike the baseline without RAG).}
\label{fig:rag-pipeline}
\end{figure}

 
To address this gap, we investigate to what extent 
external knowledge can support the generation of high-quality explanations for implicit misogynistic messages.
To this end, we propose a Retrieval-Augmented Generation (RAG) framework for detecting and explaining implicit misogyny in English and Italian in online discourse.
Figure \ref{fig:rag-pipeline} shows our pipeline.
By grounding explanations in retrieved contextual evidence, our approach aims to produce outputs that are more interpretable and useful for understanding implicit misogyny in real-world online settings, and respond to harmful content that would otherwise remain difficult to justify under platform policies.
We aim to answer the following research questions: \\
\textbf{RQ1.} Does the posting context of a comment help detect implicit misogyny?\\
\textbf{RQ2.} Does RAG improve LLMs’ ability to detect and explain implicit misogyny?\\
\textbf{RQ3.} Which knowledge source---encyclopedic, slang, or feminist literature---best supports context-grounded explanations?\\
\textbf{RQ4.} To what extent do RAG models address known failure modes?\\
Overall, RAG improves misogyny detection on \textsc{ImplicIT-Mis}, with feminist literature achieving the highest recall, but not on \textsc{SBIC+}, where the no-RAG baseline performs best. Human evaluation, however, shows that RAG is more consistently beneficial for explanation quality, producing more specific explanations that annotators prefer over the baseline in both languages.

\paragraph{Our Contributions}
i) We introduce a RAG framework for detecting and explaining implicit misogyny using multiple external knowledge sources; 
ii) We evaluate explanation quality through human annotation and diagnose where RAG mitigates known LLM weaknesses;
iii) We release our pipeline and analysis at \href{https://github.com/ivana2003/rag-implicit-mis/}{our repository}.

\section{Related Work}
Following \citet{waseem-etal-2017-understanding}, we refer to implicit misogyny as abusive language that ``does not immediately imply or denote abuse'', its ``true nature [being] often obscured by the use of ambiguous terms, sarcasm, lack of profanity or hateful terms and other means'', as opposed to explicit abuse, which is ``unambiguous in its potential to be abusive''. Implicit misogyny therefore requires inference over meanings that are not directly stated, relying on background knowledge or pragmatic inferences. This makes it challenging for NLP systems, since lexical cues that help explicit hate-speech detection may not transfer to implicit cases: for example, entity mentions can improve explicit hate-speech detection while worsening implicit hate-speech detection \cite{lin-2022-leveraging}. Recent work has addressed this challenge by integrating external knowledge into harmful-language detection, including retrieval-based models for multilingual hate speech \cite{mnassri-etal-2025-rag} and knowledge-based approaches for implicit or offensive content \cite{zhao-etal-2025-enhancing-llm, cuccarini-etal-2025-unveiling, liu2026chinese}. However, these approaches mainly target classification, leaving open whether retrieval can also support explanations of why a text is uncivil.

A related line of work focuses on explanation and reasoning for implicit hate. \citet{alajmi2026doubtconsultexpertdebate} generate structured explanations for implicit sexism through a multi-persona debate framework, while \citet{muti-etal-2024-language, muti-etal-2025-r} show that LLMs often produce implied assumptions by relying on semantic associations rather than robust reasoning. Similar challenges arise in metaphorical hate speech, where harmful meaning is conveyed through figurative language and remains difficult for LLMs \cite{zeng-etal-2025-sheeps}. At the same time, retrieval is increasingly used in moderation-oriented settings: \citet{bonaldi2026catchme} introduce a counterspeech dataset grounded in verified external knowledge. 
Together, this body of literature motivates our central question: can retrieval help models not only detect implicit misogyny, but also generate explanations grounded in the right contextual knowledge?


\section{External Knowledge Sources}
\label{sec:kb}

We construct three domain-specific knowledge bases to target different types of implicit meaning: 

i.) Wikipedia is expected to help when misogyny depends on named entities, public figures, or factual references; ii.) the slang lexicon, 
when misogyny is encoded through informal, derogatory, or Incel-specific vocabulary; and iii.) feminist literature, 
when the implicit meaning depends on theoretical concepts such as dismissal of feminism, sexual objectification and victim-blaming \cite{muti-etal-2025-r}. 


\subsection{Wikipedia Knowledge Base}
The Wikipedia knowledge base is organized into two distinct stages designed to maximize domain relevance while maintaining broad coverage. In the first stage, we filter the Wikidata dump to retain only entities directly relevant to the task domain: human entities, identified by Wikidata class Q5; and three in-domain topics, i.e., feminism (Q7252), femicide (Q1342425) and violence against women (Q1800556). 
These categories represent complementary dimensions of misogynistic discourse: its human targets, its manifestations of gender-based violence, and the feminist movements and debates that such discourse frequently opposes. We first experimented with the entire unfiltered dump, however it produced substantial noise with generic entities frequently linked to unrelated or uninformative pages. Restricting the retrieval upstream keeps the retrieval anchored to the task domain.
In the second stage, we retrieve each entity’s Wikipedia page via the Wikipedia API and index it locally using Whoosh \citep{chaput_whoosh}. For each extracted entity, we search this domain-filtered index first, falling back to the live Wikipedia Search API for broader coverage when no match is found.

\subsection{Slang Knowledge Base}


For Italian, we use the
\texttt{en.wiktionary.org} category \texttt{Italian\_slang}, comprising approximately
530 entries. 
For English, we combine three Wiktionary categories: \texttt{English\_slang}, \texttt{English\_internet\_slang}, \texttt{English\_incel\_slang}, 
yielding approximately 22,000 entries. 
Among all candidate matches for a given token, we select the definitions tagged as \texttt{vulgar}, \texttt{derogatory}, or \texttt{offensive} in the Wiktionary markup, ensuring that the offensive sense is selected when a word has multiple definitions.

\subsection{Feminist Literature Knowledge Base}
The knowledge base is built from two books: \textit{Down Girl: The Logic of Misogyny} 
\citep{manne2018down} and \textit{Gender Hate Online: Understanding the New 
Anti-Feminism} \citep{ging2019gender}. The first one provides a philosophical account of misogyny as a system of enforcement, distinguishing it from sexism and tracing how it operates through social norms and expectations. The second one offers an empirical analysis of contemporary anti-feminist movements online, covering incel communities, men's rights activism and the rhetorical strategies through which misogynistic ideology is expressed in digital spaces. Together, the two texts provide complementary angles on implicit misogyny: one theoretical and one empirical and grounded in online discourse.

\section{Datasets}

We conduct experiments on two datasets: \textsc{ImplicIT-Mis} \cite{muti-etal-2024-language} for Italian and the misogynistic subsection of \textsc{SBIC} \cite{sap-etal-2020-social} and \textsc{Implicit Hate} \cite{elsherief-etal-2021-latent} for English.

\textsc{ImplicIT-Mis} is an Italian corpus of 1,120 Facebook comments collected from women-related news articles. All instances are annotated as implicitly misogynistic. The dataset is especially relevant for our work because many comments are tied to news events, public figures, or culturally specific references, making it a useful testbed for evaluating whether contextual information improves detection and explanation in a non-English setting. 

\textsc{SBIC+} is an English dataset of 2,344 implicitly misogynistic messages constructed from filtered instances of the Social Bias Inference Corpus (SBIC) and the Implicit Hate Corpus (IHC) targeting women. Compared to \textsc{ImplicIT-Mis}, \textsc{SBIC+} is less implicit than \textsc{ImplicIT-Mis}, as it contains more slurs and explicit jokes with the structure \textit{what's the difference between women and X?}.


\section{Methodology}

We design a RAG pipeline to construct an external knowledge representation for each input sentence. 
Our pipeline applies domain filtering, entity-driven retrieval, a first entity-driven compression, semantic reranking and a second context-driven compression.
All embedding-based steps use the same multilingual sentence transformer, \texttt{paraphrase-multilingual-MiniLM-L12-v2}. In the pipeline we use \textit{input} to refer to the message to be classified together with its context when it is available (only for ImplicIT-Mis).

\paragraph{Entity Extraction and Filtering}
Named entities are initially extracted using spaCy \citep{spacy} and expanded 
with a set of rule-based strategies. First, partial person names are recombined into full names by searching the local context for adjacent capitalized tokens that plausibly belong to the same entity. Second, we identify sequences of consecutive capitalized tokens that were not tagged as entities at all by the base model. A filtering stage removes short expressions, function words, and duplicates.
Finally, we retain domain-relevant entities using a predefined keyword set: \textit{violence, femicide, rape, harassment, victim, woman, misogyny, sexism, abuse}, established prior to any inspection of the dataset to avoid overfitting on our datasets and support generalization.


\paragraph{Candidate Retrieval}
For Wikipedia, candidate documents are retrieved both from the local index and the live Wikipedia API that are subsequently merged and deduplicated.
For slang, we match input tokens against the Wiktionary slang lexicon and we retrieve the Wiktionary entry of each matched term.
For literature, we retrieve the two most similar two-sentence units from the knowledge base, provided their cosine similarity exceeds 0.6. This value has been empirically selected.

\paragraph{First Compression}
Following \citet{xu2024recomp}, we compress retrieved evidence before in-context integration.
The procedure varies by source, because retrieval units differ in length. Wikipedia 
pages are split into sentences and each sentence receives an additive score: $+3$ if it mentions one of the target entities, $+2$ if it contains a domain keyword, and $+1$ if it is among the first three sentences of the article, reflecting the editorial convention by which the lead section of a Wikipedia article should stand on its own as a concise overview of the topic \cite{wikipedia-lead}. At most 8 sentences are retained and the result is truncated to 1,500 characters since beyond that length we observed that the retrieved content started to include irrelevant information. For slang, compression operates at the level of word senses rather than sentences: among all the definitions Wiktionary lists for a matched term, we keep only those 
whose markup contains \texttt{vulgar}, \texttt{derogatory} or \texttt{offensive} tags.
For literature, we do not apply compression, as passages are indexed as two-sentence units.

\paragraph{Semantic Reranking}
Reranking is applied to Wikipedia candidates entity by entity. Each candidate page gets a score that sums two cosine similarities: (i) with the entity string and (ii) with the input.
We retain at most 3 candidates per entity, subject to at least one of the following conditions: the cosine similarity is at least 0.3 or the entity appears in the title or in the extract. The resulting retrieval field is truncated again at 4,500 characters, consistent with aggregating at most three candidates of 1,500 characters each. Slang entries are reranked by cosine similarity between the input embedding and the embedding of the word-definition pair, retaining only matches above 0.3.  
For literature, retrieval and reranking coincide since the two most similar units are selected directly by cosine similarity.

\paragraph{Second Compression}
\label{par:second-compression}
A second compression stage filters the aggregated retrieval field at sentence level, retaining only sentences whose cosine similarity to the input exceeds 0.3. It is applied to Wikipedia and literature, whose retrieved passages are still long at this point, while slang is left untouched since dictionary definitions are brief by construction.
All threshold values were selected empirically through qualitative inspection of the retrieved evidence; the selection procedure is described in Appendix~\ref{app:similarity-thresholds}.

\subsection{Retrieval Coverage}
\label{sec:coverage}
We explore retrieval coverage across datasets, which varies across knowledge sources and languages.
Table \ref{tab:coverage} shows a clear complementarity pattern: Wikipedia achieves high coverage for ImplicIT-Mis (87.4\%) but near-zero coverage for SBIC+ (3.4\%), reflecting the great presence of named entities in ImplicIT-Mis, as news often target specific people. The slang knowledge base shows the opposite asymmetry (9.7\% ImplicIT-Mis vs 57.1\% SBIC+), consistent with the derogatory vocabulary characteristic of SBIC+. Feminist literature provides more balanced coverage across both languages (29.6\% ImplicIT-Mis vs 40.8\% SBIC+).

\begin{table}[h]
\centering
\footnotesize
\begin{tabular}{lrr}
\toprule
\textbf{KB} & \textbf{IT} & \textbf{EN} \\
\midrule
Wikipedia  & 87.4\% & 3.4\%    \\
Slang      & 9.7\%  & 57.1\% \\
Literature & 29.6\% & 40.8\%  \\
\bottomrule
\end{tabular}
\caption{Retrieval coverage for each knowledge base.}
\label{tab:coverage}
\end{table}


\section{Experiments}


Following the findings of \citet{muti-etal-2024-language}, we select the best-performing model reported in their study for our experiments. Specifically, we use 
\textsc{Llama-3.1-8B-Instruct}, the most recent version available at the time of experimentation. 
We set the temperature to 0.3 to balance reproducibility and linguistic variability and we set max\_new\_tokens to 256 to accommodate the JSON output without truncation.
All the other parameters are left at default.

We design different prompts corresponding to the research questions introduced in Section~\ref{sec:introduction}. We report some considerations on prompt variations and the prompts used in Appendix \ref{app:prompts}.
Since all instances in the datasets are misogynistic, model performance is evaluated using Recall. For the Italian dataset, we evaluate prompts written both in Italian and in English.

\paragraph{Text-only baseline.}
As a baseline, we prompt the model using only the misogynistic text. This setting measures the model's ability to identify implicit misogyny from the surface text alone. 

\paragraph{Context setting (+ headline).}
Implicit messages require additional context in order to be fully understood. The \textsc{ImplicIT-Mis} dataset allows access to the headlines of the news articles under which the Facebook comments were posted. We therefore augment the context by passing the model both the news headline and the related comment. This setting directly addresses RQ1, 
testing whether contextual information improves implicit misogyny detection. 

\paragraph{RAG setting.}
For each input message, we retrieve evidence from one of our knowledge bases (Wikipedia, the slang lexicon, and feminist literature). The retrieved evidence is then appended to the prompt as additional context. 
This setting directly addresses RQ2, assessing the extent to which RAG can be used to improve implicit misogyny detection and generate better explanations.




\subsection{Qualitative Evaluation of Generated Explanations}

We assess through a qualitative analysis whether the generated explanations for the misogynistic messages are of a better quality than those obtained from the baselines. We define ``better'' along two dimensions. The first concerns the \textit{faithfulness} of the generated explanation with respect to the target message. In particular, faithfulness measures the extent to which an explanation is supported by the input text and, when applicable, by the retrieved evidence. Faithfulness is annotated on a three-point scale: 2 when the explanation identifies the implicit misogynistic assumption and is clearly supported by the retrieved evidence, 1 when it captures part of the intended meaning but is incomplete, vague or weakly connected to the evidence and 0 when it is unsupported, contradicts the text or invents content. 
Faithfulness scores are computed as follows: each explanation is scored by two annotators, so for each knowledge base we 
obtain $N$ (twice the number of available explanations) scores that sum up to a ''Sum'' variable. 
Faithfulness is the ''Sum'' variable divided by $N$ and therefore ranges 
in $[0,2]$. 

The second dimension concerns \textit{preference}, 
which is more subjective, as annotators had to select which explanation they preferred. 
They may select \textit{Tie} when no explanation is clearly best or \textit{None} when all are incorrect.
Human preference reflects how many times, across both annotators, a given system was selected as best, out of the total number of explanations available. Normalizing by the availability of the retrieved evidence is necessary because texts do not have an equal number of retrieved evidence.

Explanations were randomized and system labels hidden.
Annotation was carried out independently by two annotators, both authors of this paper, who are experts in misogynistic language. On faithfulness, they reached a Cohen's $\kappa$ agreement of 0.505 for Italian and 0.454 for English. On preference, they reached $0.333$ for English and $0.286$ for Italian. The preference score was lower as it reflects a more subjective judgment. After the initial annotation, the annotators discussed the disagreement cases and clarified how the guidelines should be interpreted. They then independently re-annotated the disagreement cases, reaching $0.925$ agreement for Italian and $0.702$ for English. The complete annotation guidelines, along with more information on the inter-annotator agreement and annotators are reported in Appendix~\ref{app:guidelines}. 


\section{Detection Results}
\label{sec:results}
\paragraph{RQ1: Posting Context Improves Misogyny Detection}
Table~\ref{tab:rq1_context} reports results on \textsc{ImplicIT-Mis}, the only dataset that includes posting context. We compare a text-only baseline with a context-enriched variant that also incorporates the post headline. 
As the figures show, enriching the original message with its posting context consistently improves performance regardless of the language used in the prompt. The largest improvement is observed when prompting the model in English (+0.120), while Italian prompts also slightly benefit from contextual information (+0.027). 


\begin{table}[t]
\centering
\footnotesize
\begin{tabular}{llr}
\toprule
\textbf{Dataset} & \textbf{Prompt setting} & \textbf{Recall}\\
\midrule
\multirow{4}{*}{\textsc{Implicit-MIS}} 
& \tt text-only - IT & 0.838  \\
& \tt text-only - EN & 0.771 \\
& \tt + headline - IT & 0.865 \\
& \tt + headline - EN & \bf 0.891 \\
\bottomrule
\end{tabular}
\caption{Impact of context (headline) and prompt language (IT \textit{vs}. EN) on misogyny detection.
}
\label{tab:rq1_context}
\end{table}

\paragraph{RQ2: Retrieval Improves Detection on \textsc{ImplicIT-Mis}, but not on \textsc{SBIC+}}
Table~\ref{tab:rq2_rag} presents the results for detection when incorporating external knowledge via RAG.
On \textsc{ImplicIT-Mis}, 
RAG generally improves performance over the best baseline (\texttt{+ headline}) (0.891), particularly when prompts are written in Italian. The best result is obtained with literature-based retrieval (0.923), followed by slang-based retrieval (0.916). In contrast, on \textsc{SBIC+}, 
none of the retrieval sources outperform the baseline (0.966), suggesting that external knowledge achieves higher performance on the Italian dataset, where cultural references may require additional background information.
Since in our experimental setup language and dataset vary jointly, these results should not be interpreted as evidence of an intrinsic difference between Italian and English. \textsc{ImplicIT-Mis} and \textsc{SBIC+} present strong differences concerning their degree of implicitness, retrieval coverage and the type of external knowledge they require, all of which may contribute to the observed difference in RAG effectiveness.
\begin{table}[t]
\centering
\footnotesize
\begin{tabular}{lllr}
\toprule
\textbf{Dataset} & \textbf{System} & \textbf{Prompt} & \textbf{Recall} \\
\midrule
\multirow{7}{*}{\textsc{Implicit-MIS}}
& \tt + headline & EN & 0.891 \\
& \tt Wikipedia & IT & 0.881 \\
& \tt Wikipedia & EN & 0.853 \\
& \tt Slang & IT & 0.916 \\
& \tt Slang  & EN & 0.895 \\
& \tt Literature & IT & \textbf{0.923} \\
& \tt Literature & EN & 0.666 \\
\midrule
\multirow{4}{*}{\textsc{SBIC+}}
& \tt baseline & EN & \textbf{0.966} \\
& \tt Wikipedia & EN & 0.962 \\
& \tt Slang  & EN & 0.806 \\
& \tt Literature & EN & 0.723 \\
\bottomrule
\end{tabular}
\caption{Comparison between baselines (\texttt{+ headline} for \textsc{ImplicIT-Mis} and \texttt{baseline} for \textsc{SBIC+}) and different retrieval sources for misogyny detection. 
}
\label{tab:rq2_rag}
\end{table}

\section{Explanation Quality}
\paragraph{Human Evaluation.} We sample 100 Italian and 100 English misogynistic texts.
We give priority to texts with a higher number of retrieved contexts. For each instance, we compare explanations produced by the baseline (no-RAG) system and by the three retrieval settings. Only instances for which retrieval was successfully performed for at least one knowledge source are included in the evaluation. In total, we have 331 explanations for Italian (100 baseline, 97 Wikipedia, 47 slang, 87 literature) and 303 explanations for English (100 baseline, 18 Wikipedia, 92 slang, 93 literature). 



\begin{table}[t]
\centering
\footnotesize
\setlength{\tabcolsep}{1.3pt}
\begin{tabular}{lcccccccc}
\toprule
& \multicolumn{4}{c}{\bf \textsc{Implicit-MIS}} & \multicolumn{4}{c}{\bf \textsc{SBIC+}} \\
\cmidrule(lr){2-5} \cmidrule(lr){6-9}
\textbf{System} & \textbf{N} & \textbf{Sum} & \textbf{Mean} & \textbf{Std.}
                & \textbf{N} & \textbf{Sum} & \textbf{Mean} & \textbf{Std.} \\
\midrule
\tt Wikipedia   & 194 & 128 & 0.659 & 0.774 & 36  & 39  & 1.083 & 0.841 \\
\tt Slang       & 94  & 72  & 0.765 & 0.835 & 184 & 240 & 1.304 & 0.833 \\
\tt Literature  & 174 & 161 & \textbf{0.925} & 0.826 & 186 & 253 & \textbf{1.360} & 0.802 \\
\tt no RAG    & 200 & 118 & 0.590 & 0.745 & 200 & 269 & 1.345 & 0.774 \\
\bottomrule
\end{tabular}
\caption{Faithfulness scores for generated explanations, aggregated across two annotators and 100 samples per language. \textbf{N} indicates the number of scores, i.e. twice the number of evaluated explanations, \textbf{Sum} is the total faithfulness score, and \textbf{Mean} reports the average faithfulness score.}
\label{tab:faithfulness-scores}
\end{table}

\paragraph{RAG Explanations Are More Faithful in Italian.} Table~\ref{tab:faithfulness-scores} shows the faithfulness scores. 
For \textsc{ImplicIT-Mis}, all retrieval settings obtain higher mean faithfulness than the baseline. In particular, Wikipedia reaches 0.659, the slang lexicon 0.765, and feminist literature 0.925, compared to 0.590 for the baseline. The aggregated RAG score is also higher than the baseline (0.781 vs. 0.590), suggesting that \textbf{RAG 
generally helps produce explanations that are better supported by the input and retrieved evidence.} For \textsc{SBIC+}, 
the pattern is more mixed. Literature-based retrieval achieves the highest mean faithfulness score (1.360), slightly above the baseline (1.345), while slang retrieval is close to the baseline (1.304) and Wikipedia performs lower (1.083). Overall, the aggregated RAG score in English is slightly below the baseline (1.310 vs. 1.345), indicating that RAG is under-performing.

We further test whether these differences are 
statistically significant. On \textsc{ImplicIT-Mis}, RAG explanations are significantly more faithful than the baseline for both annotators, with a higher average faithfulness score (mean RAG = 0.788 vs. baseline = 0.590; $\Delta = +0.198$). This difference is significant (Wilcoxon test; $p < 0.0001$), with a small positive effect size (Cohen's $d = 0.272$). On \textsc{SBIC+}, however, we do not observe a significant difference between RAG and the baseline. This indicates that the no-RAG model already produces faithful explanations in English, leaving less room for retrieval to improve faithfulness.

\paragraph{RAG Explanations Are Preferred over the Baseline.}
Table~\ref{tab:rag-vs-baseline} shows that \textbf{RAG-based explanations are preferred more often than the no-RAG baseline in both languages}. In Italian, RAG explanations receive 121 preferences compared to 51 for the baseline; in English, they receive 154 preferences compared to 95. 

\begin{table}[t]
\centering
\footnotesize
\setlength{\tabcolsep}{1.3pt}
\begin{tabular}{lrrr}
\toprule
\textbf{Dataset} & \textbf{N. of texts} & \textbf{RAG} & \textbf{Baseline} \\
\midrule
\textsc{Implicit-MIS} & 200 & 121 & 51 \\
\textsc{SBIC+} & 200 & 154 & 95 \\
\bottomrule
\end{tabular}
\caption{Aggregate preference across two annotators. A text can count in both columns in case of ties or disagreements.}
\label{tab:rag-vs-baseline}
\end{table}

\begin{table}[t]
\centering
\footnotesize
\setlength{\tabcolsep}{1.3pt}
\begin{tabular}{lrrrr}
\toprule
\textbf{Dataset} & \textbf{RAG Wins} & \textbf{Baseline Wins} & \textbf{Tie} & \textbf{Neither} \\
\midrule
\textsc{Implicit-MIS} & 88 & 18 & 33 & 61 \\
\textsc{SBIC+} & 86 & 27 & 68 & 19 \\
\bottomrule
\end{tabular}
\caption{Mutually exclusive breakdown of the comparison in Table~\ref{tab:rag-vs-baseline}. \textit{Neither} indicates that no explanation was selected.}
\label{tab:rag-baseline-mutual}
\end{table}

Table~\ref{tab:rag-baseline-mutual} provides a mutually exclusive comparison between RAG-based and no-RAG explanations. In both languages, RAG is substantially more often preferred 
than the baseline: 88 vs. 18 cases in Italian and 86 vs. 27 cases in English. English also shows a higher number of ties between RAG and baseline, suggesting that baseline explanations are more competitive in this setting. By contrast, \textsc{ImplicIT-Mis} has more cases in which neither system is preferred, indicating that explaining implicit misogyny in Italian remains challenging for LLMs, even with retrieval.

\paragraph{RQ3: Feminist Literature Provides the Most Useful Explanatory Context}

We analyze which retrieval source contributes most to this improvement in the generation of the explanations. Table~\ref{tab:human-eval} reports the preferences by 
knowledge source, normalized by the number of cases in which each source was available. 
Results show that retrieval improves explanation quality in both languages, but the effect depends on the retrieval source. 
For \textsc{ImplicIT-Mis}, literature-based retrieval is preferred most often, with the highest normalized score (0.408), outperforming the no-RAG baseline (0.215), Wikipedia (0.237), and the slang lexicon (0.255). 
For \textsc{SBIC+}, literature also achieves the highest normalized preference score (0.538), followed by the slang lexicon (0.500), both outperforming the no-RAG baseline (0.460). By contrast, Wikipedia performs below the baseline in English (0.361), likely reflecting its limited coverage and lower relevance for explaining implicit misogyny in this dataset. Overall, these results suggest that retrieval can improve explanation quality when the retrieved evidence provides relevant social or linguistic context, with feminist literature emerging as the most consistently useful source across languages.

Examples of texts where RAG systems support the explanations are presented in Table \ref{tab:rag-failure-modes-examples}.
\paragraph{Wikipedia can make explanations more specific.}
Wikipedia helps in Italian when the implicit meaning depends on a cultural reference. In Example (1) the baseline produces a wrong assumption, while the retrieved context on the femicide case and on Elena Cecchettin's public denunciation of patriarchy lets RAG identify the more specific assumption that women who speak publicly about their suffering are not dignified.

\paragraph{Feminist literature supports logical inference}
Feminist literature helps when the implicit meaning requires reasoning grounded in ideological context. 
In Example (2), the retrieved text characterizes patriarchy as a hegemonic social order, allowing the model to infer that the comment denies patriarchy as a structural cause of femicide. This additional context helps the model reconstruct the implicit assumption that \textit{``patriarchy is considered positive and desirable.''}
Literature also helps with irony: in Example (3) the baseline reads \textit{``female superiority''} as a neutral statement about women's autonomy, while the retrieved frame on how female excellence can become polarizing lets RAG recover the ironic reading.

\paragraph{Slang retrieval can recover community-specific derogatory meanings and figurative language.} 
Slang retrieval is especially useful in English, when the misogynistic implication depends on community-specific vocabulary, like Incel, which use figurative language to refer to women. In Example (4), the baseline treats \textit{roasties} as something women wear. The retrieved incel-slang definition instead identifies \textit{roastie} as a derogatory term for women, allowing RAG to interpret the comment as relying on the metaphor of makeup as a ``mask''.

\begin{table}[!t]
\centering
\footnotesize
\begin{tabular}{llrr}
\toprule
\textbf{Dataset} & \textbf{System} & \textbf{Preferred} & \textbf{Normalized} \\
\midrule
\multirow{4}{*}{\textsc{Implicit-MIS}}
& \texttt{Wikipedia} & 46/194 & 0.237 \\
& \texttt{Slang} & 24/94 & 0.255 \\
& \texttt{Literature} & 71/174 & \textbf{0.408} \\
& \texttt{+ headline} & 43/200 & 0.215 \\
\midrule
\multirow{4}{*}{\textsc{SBIC+}}
& \texttt{Wikipedia} & 13/36 & 0.361 \\
& \texttt{Slang} & 92/184 & 0.500 \\
& \texttt{Literature} & 100/186 & \textbf{0.538} \\
& \texttt{baseline} & 92/200 & 0.460 \\
\bottomrule
\end{tabular}%

\caption{Preference by knowledge source. \textbf{Preferred} reports how often each system's explanation was preferred, over the number of explanations produced.}
\label{tab:human-eval}
\end{table}

\begin{table*}[!t]
\centering
\scriptsize
\setlength{\tabcolsep}{3.5pt}

\begin{tabular}{p{0.03\linewidth}p{0.09\linewidth}p{0.07\linewidth}p{0.35\linewidth}p{0.20\linewidth}p{0.18\linewidth}}

\toprule
\textbf{\#} & \textbf{Failure mode} & \textbf{RAG source} & \textbf{Input text} & \textbf{Baseline explanation} & \textbf{RAG explanation} \\
\midrule

(1) & \textbf{Lack of \newline Reference}
& Wikipedia
& \textit{``Dignified is silence, not the spectacularization of an announced tragedy. You are killing again the other murdered women; remain in the dignified silence of relatives. But the show must go on, right?''} Context: a post suggesting Elena Cecchettin as ``woman of the year'' for speaking up about femicides.
& Women are exploited to attract audiences and attention.
& Women who speak publicly about their suffering are not dignified. \\
\midrule

(2) & \textbf{Wrong \newline inference}
& Literature
& \textit{``If only patriarchy existed''} (translated from IT), in response to a headline about Elena Cecchettin's speech on femicide.
& Patriarchy protects women from situations such as femicide.
& Patriarchy is considered positive and desirable. \\

\midrule
(3) & \textbf{Sarcasm /\allowbreak Irony}
& Literature
& \textit{``The female superiority (translated from IT)''}
& Women have greater awareness and autonomy than their parents.
& Female superiority is viewed with suspicion or as a negative idea. \\

\midrule

(4) & \textbf{Figurative language}
& Slang
& \textit{``"makeup
tell me about roasties, why do they wear the mask?"''}
& women (specifically those who wear 'roasties' or possibly those who wear makeup) are hiding or disguising their true appearance.
& women wear makeup to hide their appearance. \\

\bottomrule
\end{tabular}
\caption{Examples of high-quality RAG explanations addressing known failure modes in implicit misogyny.}
\label{tab:rag-failure-modes-examples}
\end{table*}

\paragraph{RQ4: RAG Address LLM's Failure Modes, But Still Falls Short On Reasoning}
To answer RQ4, we conduct a diagnostic analysis combining quantitative counts of resolved failure modes with a qualitative inspection of representative examples.
For this analysis, we select the RAG sentences that have received a faithfulness score of 2 and the baseline is different from 2, or that at least one RAG was preferred over the baseline. We are left with 35 sentences for IT and 36 for EN.
We consider the taxonomy of LLM failure modes in implicit misogyny proposed by \citet{muti-etal-2025-r}. The taxonomy is composed by seven classes of errors: sarcasm and irony, metaphorical or figurative language, wrong translations (for Italian only), opposite intention, denial of misogyny, lack of reference, and wrong inference. 
We first identify the sentences presenting these problems, which are 20 for IT and 12 for EN.
For each text presenting one of these problems, we check whether the RAG explanation addresses one of these known challenges; that is, whether it correctly identifies the implicit mechanism that LLMs typically fail to capture. 
For Italian, 21 issues are solved by any of the RAG explanations, which are divided as follows: wrong inference (9 occurrences); opposite intention (6), 
denial of misogyny (3), sarcasm/irony (2), figurative/metaphorical language (1).
For English, 9 issues are solved by any of the RAG explanations, which are divided as follows: wrong inference (4), sarcasm/irony (4), lack of reference (1).

Table \ref{tab:rag-failure-modes-examples} shows examples of four failure modes solved by the RAG systems: lack of reference, wrong inference, sarcasm, and figurative language.



\paragraph{Explanations are still driven by lexical associations.}
Although RAG explanations are often preferred by annotators, preference does not necessarily imply correct reasoning. Recent work on prediction--explanation consistency shows that LLM-generated explanations can be plausible while failing to consistently support the model's prediction \cite{zhao-iii-2025-necessary}. This is relevant to our findings: in some cases, RAG models correctly predict misogyny, but the generated explanation is only lexically related to the input and does not provide the pragmatic support needed for the explanation. 
In English, in Example \textit{``this roastie thinks she and other women can be fa hahahahahaha''}, the slang retrieval correctly identifies \textit{roastie} as incel terminology, but the generated explanation does not fully recover the implied assumption. The comment mocks the idea that women can be \textit{forever alone}, since incel ideology assumes that women can always obtain sexual or romantic attention. Instead, the RAG explanation focuses on promiscuity, capturing part of the derogatory framing but missing the central implication of the joke.
In Italian, the comment \textit{``For men, it is enough to say one wrong word at the wrong time to get beaten up''} responds to a post where a man sarcastically describes a woman covering her body at night for safety: \textit{``Oversized sweatshirt, sweatpants, completely covered! Not one boob out! That is not life.''} The slang retriever returns an entry for \textit{uscirle}---literally \textit{``to get them out''}---meaning to expose one's breasts. This retrieval is triggered by lexical overlap, but it does not explain the misogynistic implication of the comment, which minimizes women's fear of male violence by reframing men as the real victims. 

\section{Conclusion}
We investigate whether RAG supports the detection and explanation of implicit misogyny in English and Italian. Results show that RAG improves detection on \textsc{ImplicIT-Mis}, especially with feminist literature and slang knowledge, but not on \textsc{SBIC+}. 
Our manual analysis suggests that these differences are inherent to the datasets: \textsc{SBIC+} requires less external knowledge than \textsc{ImplicIT-Mis}. 
We also show that 
RAG-based explanations are preferred over baseline explanations more often in both languages, with feminist literature emerging as the most effective source of contextual evidence. 
However, some explanations still rely on shallow lexical associations rather than genuine pragmatic reasoning.

\section*{Limitations}
This work has some limitations. First, while both datasets are designed around implicit language, we cannot guarantee that all instances in \textsc{SBIC+} are genuinely implicit: often, English examples are short, with sexually explicit jokes, or contain explicit slurs, which affect considerations on the usefulness of retrieval. 
Second, human evaluation is conducted on a subsample of 100 instances per language, so the preference and faithfulness results should be interpreted as diagnostic rather than exhaustive. 
Third, retrieval coverage varies substantially across knowledge sources and languages, which affects how directly source-level results can be compared. Fourth, retrieved evidence is not always pragmatically relevant: in some cases, retrieval is driven by lexical overlap rather than by the implicit reasoning needed to explain misogyny. 
Fifth, annotations are carried out by the authors of the paper. However, i) they are expert in recognizing misogynistic language; ii) the explanations were anonymized, so that the annotators did not know in advance which system explanation they were rating. Finally, our main experiments use a single 8B model. Although Appendix~\ref{app:model-sizes} reports additional results with smaller models from the same family, future work should test whether these findings generalize across model families and larger model scales.

\section*{Ethical Considerations}
\paragraph{Copyrighted books.} In accordance with the applicable European text-and-data-mining exception for scientific research, the texts and derived index were stored securely and used only by the research team. Because the source texts are copyrighted, we do not release the texts or derived index.

\paragraph{Misogynistic examples.} The misogynistic examples reported in this paper and in the released resources do not reflect the positions or beliefs of the authors and are included solely for the purpose of presenting and analyzing the data. Although we recognize that presenting hateful content may unintentionally expose readers to offensive language, we have chosen not to obscure or censor the examples. Preserving the original wording is important for ensuring transparency, reproducibility, and accurate analysis, as even minor modifications can alter the linguistic characteristics that are central to this research.

\paragraph{Declaration on the use of AI.} The authors used AI tools in order to: Grammar and spelling check; Paraphrasing. After using this tool, the authors reviewed and edited the content as needed and take full responsibility for the publication’s content.

\section*{Acknowledgments}
Arianna Muti's and Debora Nozza's research is supported by the European Research Council (ERC) under the European Union’s Horizon 2020 research and innovation program (grant agreement No. 101116095, PERSONAE). Arianna Muti, Ivana Crescenzi and Debora Nozza are members of the MilaNLP group and the Data and Marketing Insights Unit of the Bocconi Institute for Data Science and Analysis.

\bibliography{custom, anthology-1, anthology-2}

\appendix
\section{Prompts}
\label{app:prompts}
We report the prompt templates used in our experiments. The placeholder \texttt{\{text\}} indicates the input text received, \texttt{\{context\}} indicates the original contextual information provided with Italian instances, and \texttt{\{retrieved\_text\}} indicates the external knowledge retrieved by the RAG pipeline. Prompt formulations were defined iteratively during development: multiple variants were qualitatively compared and the versions producing most consistent and well-formatted JSON outputs were retained.

\subsection{English baseline}
\label{app: prompt-en-baseline}
\begin{prompt}
You are a classifier for implicit misogyny.
What are the implied assumptions of the text {text}?
Based on the implied assumptions, is the text misogynous?
Return ONLY valid JSON.

Format:
{
  "text": "string",
  "implied_assumption": "string",
  "answer": "YES or NO"
}

Follow the examples.

[
  {
    "text": "we shouldn't lower our standards just to hire more women.",
    "implied_assumption": "women are less qualified than men.",
    "answer": "YES"
  },
  {
    "text": "Women shouldn't talk about football.",
    "implied_assumption": "women cannot understand football.",
    "answer": "YES"
  },
  {
    "text": "She must be on her period.",
    "implied_assumption": "women are irrational or overly emotional.",
    "answer": "YES"
  }
]

Text: {text}
\end{prompt}

\subsection{English Wikipedia}
\label{app: prompt-en-wiki}
\begin{prompt}
You are a classifier for implicit misogyny.
What are the implied assumptions of the text {text}?
Based on the implied assumptions and the retrieved_text provided, is the text misogynous?
The value of "answer" must be only YES or NO.
Return ONLY valid JSON.

Format:
{
  "text": "string",
  "retrieved_text": "string",
  "implied_assumption": "string",
  "answer": "YES or NO"
}

Follow the examples.

[
  {
    "text": "if you watch cinderella backwards its about a woman being put in her place.",
    "retrieved_text": "### Cinderella
The earliest known version of the Cinderella story is usually considered to be the Greek story of Rhodopis, as described by the scholar Strabo sometime between 7 BC and AD 23, about a Greek slave girl who marries the king of Egypt. ==== Le Fresne ====

The twelfth-century lai of Le Fresne (\"The Ash-Tree Girl\"), retold by Marie de France, is a variant of the \"Cinderella\" story in which a wealthy noblewoman abandons her infant daughter at the base of an ash tree outside a nunnery with a ring and brocade as tokens of her identity. ==== Kongjwi and Patjwi ====
Originating from Korea, Kongjwi and Patjwi is a tale similar to Disney's Cinderella, with two distinguishing characteristics: the degree of violence and the plot's continuance past the marriage to the prince charming. A notable difference from other versions of Cinderella is that Kongjwi avenges her death with her own determination and willpower. The violent degree of the punishments stems from the increased violence (starvation, beating, betrayal, and ultimately murder) that Kongjwi suffered compared to the abuses other versions went through. Cinderella bears the abuse patiently and does not tell her father, who would have scolded her. The prince has become even more infatuated with the mysterious woman at the ball, and Cinderella in turn becomes so enchanted by him she loses track of time and leaves only at the final stroke of midnight, losing one of her glass slippers on the steps of the palace in her haste.",
    "implied_assumption": "women should be treated like Cinderella, implying that they need to be put in their place",
    "answer": "YES"
  },
  {
    "text": "Women claim they don't want to be sexualized, yet they dress exactly like Sydney Sweeney. You can’t have it both ways.",
    "retrieved_text": "Sydney Bernice Sweeney (born September 12, 1997)[1] is an American actress. She gained early recognition for her roles in Everything Sucks!, The Handmaid's Tale, and Sharp Objects in 2018. She received wider acclaim for her performances in the drama series Euphoria (2019–present) and the first season of the anthology series The White Lotus (2021), both of which earned her nominations for Primetime Emmy Awards.[2][3]

In film, Sweeney garnered critical acclaim for her performances in the drama film Reality (2023) and for her portrayal of professional boxer Christy Martin in the biopic Christy (2025), and has also appeared in the box office hits Anyone but You (2023) and The Housemaid (2025). Her other film credits include Once Upon a Time in Hollywood (2019), Madame Web (2024), and Immaculate (2024).",
    "implied_assumption": "Dressing like Sydney Sweeney, in a provocative way, is an invitation to be sexualized.",
    "answer": "YES"
  }
]

Text: {text}
Retrieved_text: {retrieved_text}
\end{prompt}

\subsection{English Slang}
\label{app: prompt-en-slang}
\begin{prompt}
You are a classifier for implicit misogyny.

Tasks:
1. You are given a text and a set of retrieved entries from Wiktionary containing English online slang definitions.
2. Identify the implied assumption in the text.
3. Based on the implied assumption, classify the text as misogynous or not. Answer only with YES or NO.

Return ONLY valid JSON.

Example:

Text: "FGM of beef curtains
When roasties get plastic surgery to hide the beef curtain genetics, isn't it femoid genital mutilation?"
Slang: "roastie"
Definition: "(Internet slang, derogatory) A misogynistic term referring to women."
Implied assumption: "a women's vagina looks like beef curtains and it is disgusting"
Answer: YES

Now process the following:

Text: {text}
Wiktionary entries:
{retrieved_text}
\end{prompt}

\subsection{English Literature}
\label{app: prompt-en-literature}
\begin{prompt}
You are a classifier for implicit misogyny.

Tasks:
1. You are given a text and a set of retrieved entries from gender studies books.
2. Identify the implied assumption in the text.
3. Based on the implied assumption and the literature, if relevant, classify the text as misogynous or not. Answer only with YES or NO.

Return ONLY valid JSON.

Examples:

Text: "if women don't want to be treated like sex objects they shouldn't dress like one."
Literature: misogynist hostility can be anything that is suitable to serve a punitive, deterrent, or warning function.
Implied assumption: women who violate norms of modesty deserve degrading treatment as a consequence
Answer: YES

Text: "there is no such thing as rape culture in the western civilisation, feminists are women sexually frustrated. black."
Literature: Western woman has become the new 'white man's burden'" because of the demands placed on men by feminists
Implied assumption: feminists are sexually frustrated subjects who exaggerate claims of rape, recasting feminist demands as an unjust burden imposed on men.
Answer: YES

Now process the following:

Text: {text}
Literature retrieved::
{retrieved_text}
\end{prompt}

\subsection{Italian baseline (+ headline) IT}
\label{prompt-it-it-baseline}
\begin{prompt}
Sei un classificatore di misoginia implicita.
Quali sono gli assunti impliciti del testo?
Sulla base degli assunti impliciti, il testo è misogino?
Rispondere solo con SI o NO.
Restituisci SOLO JSON valido.
Rispondi come negli esempi.

Testo: non dovremmo abbassare i nostri standard solo per assumere più donne.
Contesto: Assunzione agevolata di donne disoccupate o in specifici settori: esonero dai contributi INPS per due anni sui contratti a tempo indeterminato stipulati dal 1° settembre 2024.
Assunto implicito: le donne sono meno qualificate degli uomini.
Risposta: SI.

Testo: le donne che parlano di calcio non si possono sentire.
Contesto: Per la prima volta una donna è stata scelta come prima voce telecronista per i Mondiali 2026.
Assunto implicito: le donne non capiscono niente di calcio, è una cosa da maschi.
Risposta: SI.

Testo: Dopo aver visto per sbaglio 5 minuti di Uomini e Donne ho deciso che odierò le donne.
Contesto: "Ora basta, siete ridicoli" Uomini e Donne, durissimo sfogo dell'ex tronista Ida Platano: il motivo? Davvero clamoroso.
Assunto implicito: Le donne che vanno alla trasmissione Uomini e Donne sono stupide.
Risposta: SI.

Testo: {text}
Contesto: {context}
Assunto implicito:
Risposta:
\end{prompt}

\subsection{Italian baseline (text-only) IT}
\label{prompt-it-it-text-only}
\begin{prompt}
Sei un classificatore di misoginia implicita.
Quali sono gli assunti impliciti del testo?
Sulla base degli assunti impliciti, il testo è misogino?
Rispondere solo con SÌ o NO.
Restituisci SOLO JSON valido.
Rispondi come negli esempi.

Testo: non dovremmo abbassare i nostri standard solo per assumere più donne.
Assunto implicito: le donne sono meno qualificate degli uomini.
Risposta: SÌ.

Testo: le donne che parlano di calcio non si possono sentire.
Assunto implicito: le donne non capiscono niente di calcio, è una cosa da maschi.
Risposta: SÌ.

Testo: Dopo aver visto (per sbaglio) 5 minuti di Uomini e Donne ho deciso che odierò le donne.
Assunto implicito: Le donne che vanno alla trasmissione Uomini e Donne sono stupide.
Risposta: SÌ.

Testo: {text}
Assunto implicito:
Risposta:
\end{prompt}

\subsection{Italian Wikipedia IT}
\label{prompt-it-it-wiki}
\begin{prompt}
Quali sono gli assunti impliciti del testo? Sulla base dei presupposti impliciti, il testo è misogino? Rispondere solo con SÌ o NO. Considera anche la conoscenza fornita se presente. Seguire l'esempio. Restituisci SOLO JSON valido.

Testo: non dovremmo abbassare i nostri standard solo per assumere più donne.
Contesto: Assunzione agevolata di donne disoccupate o in specifici settori: esonero dai contributi INPS per due anni sui contratti a tempo indeterminato stipulati dal 1° settembre 2024.
Assunto implicito: le donne sono meno qualificate degli uomini.
Risposta: SÌ.

Testo: le donne che parlano di calcio non si possono sentire.
Contesto: Per la prima volta una donna è stata scelta come prima voce telecronista per i Mondiali 2026. 
Assunto implicito: le donne non capiscono niente di calcio, è una cosa da maschi.
Risposta: SÌ.

Testo: Dopo aver visto (per sbaglio) 5 minuti di Uomini e Donne ho deciso che odierò le donne.
Contesto: "Ora basta, siete ridicoli" Uomini e Donne, durissimo sfogo dell'ex tronista Ida Platano: il motivo? Davvero clamoroso
Assunto implicito: Le donne che vanno alla trasmissione Uomini e Donne sono stupide.
Risposta: SÌ.

Testo: {text}
Contesto: {contesto}
Conoscenza: {retrieved_text}
Presupposto:
\end{prompt}

\subsection{Italian Slang IT}
\label{prompt-it-it-slang}
\begin{prompt}
Sei un classificatore di misoginia implicita.

Compiti:
1. Ti viene fornito un testo e un insieme di voci recuperate da Wiktionary contenenti definizioni di slang online in italiano.
2. Identifica l'assunto implicito nel testo.
3. Sulla base dell'assunto implicito, classifica il testo come misogino o meno. Rispondi solo con SI o NO.

Restituisci SOLO JSON valido.

Esempio:

Testo: "Avrà mostrato le menne per ottenere la posizione"
Contesto: "Christina Koch, prima donna verso la Luna: chi è e cosa farà su Artemis II"
Slang: "menne"
Definizione: "(dialectal slang) Women's breast"
Assunto implicito: "Christina Koch ha ottenuto la posizione sfruttando il proprio corpo, piuttosto che grazie alle sue competenze e qualifiche professionali."
Risposta: SI

Ora processa il seguente input:

Testo: {text}
Contesto: {Contesto}

Voci Wiktionary:
{retrieved_text}
\end{prompt}

\subsection{Italian Literature IT}
\label{prompt-it-it-literature}
\begin{prompt}
Sei un classificatore di misoginia implicita.

Compiti:
1. Ti viene fornito un testo e delle frasi recuperate dalla letteratura di studi di genere.
2. Identifica l'assunto implicito nel testo.
3. Sulla base dell'assunto implicito e, se rilevante, della letteratura, classifica il testo come misogino o meno. Rispondi solo con SI o NO.

Restituisci SOLO JSON valido.

Esempi:

Contesto: La festa in piscina finisce in stupro, notte horror per due 20enni. «Mi sono svegliata con un ragazzo sopra di me»
Testo: Sinceramente non mi fa pena alla fine della storia sempre alcol e droga (non in questo caso) questi giovani devono trovare altre strade per divertirsi.
Letteratura: Codes applied included the presence or absence of physical injury, perpetrator's use of a weapon or force, the circumstances of the alleged assault (such as location, presence of drugs or alcohol) and the accuser's relationship to the alleged perpetrator (stranger, acquaintance, friend, boyfriend).
Assunto implicito: Le donne che vengono stuprate vengono colpevolizzate se usano alcol o droghe
Risposta: SI

Contesto: Femminicidi, un corteo e uno sciopero alla islandese. Ecco perché i social si stanno tingendo di rosso.
Testo: Brutta storia essere aride e sterili femministe.
Letteratura: feminists that refuse to accept unjust societal conditions that include violence, harassment, silencing and other daily tactics of marginalization are depicted by different publics, from mainstream media to anti-feminist counterpublics, as disruptive and full of rage.
Assunto implicito: Le femministe sono considerate isteriche e rabbiose.
Risposta: SI

Ora processa il seguente input:

Text: {text}
Context.{Contesto}
Letteratura: {retrieved_text}
\end{prompt}

\subsection{Italian baseline (+ headline) EN}
\label{prompt-it-en-baseline}
\begin{prompt}
You are a classifier for implicit misogyny.
What are the implied assumptions of the text {text}?
Based on the implied assumptions and the context provided, is the text misogynous?
Answer only with SI or NO.
Return ONLY valid JSON.
Follow the examples.

Text: non dovremmo abbassare i nostri standard solo per assumere più donne.
Context: Assunzione agevolata di donne disoccupate o in specifici settori: esonero dai contributi INPS per due anni sui contratti a tempo indeterminato stipulati dal 1° settembre 2024.
Assumption: le donne sono meno qualificate degli uomini.
Answer: SI.

Text: le donne che parlano di calcio non si possono sentire.
Context: Per la prima volta una donna è stata scelta come prima voce telecronista per i Mondiali 2026.
Assumption: le donne non capiscono niente di calcio, è una cosa da maschi.
Answer: SI.

Text: Dopo aver visto per sbaglio 5 minuti di Uomini e Donne ho deciso che odierò le donne.
Context: "Ora basta, siete ridicoli" Uomini e Donne, durissimo sfogo dell'ex tronista Ida Platano: il motivo? Davvero clamoroso.
Assumption: Le donne che vanno alla trasmissione Uomini e Donne sono stupide.
Answer: SI.

Text: {text}
Context: {context}
Assumption:
Answer:
\end{prompt}

\subsection{Italian baseline (text-only) EN}
\label{prompt-it-en-text-only}
\begin{prompt}
You are a classifier for implicit misogyny.
What are the implied assumptions of the text {text}?
Based on the implied assumptions, is the text misogynous?
Answer only with YES or NO.
Return ONLY valid JSON.
Follow the examples.

Text: non dovremmo abbassare i nostri standard solo per assumere più donne.
Assumption: le donne sono meno qualificate degli uomini.
Answer: SÌ.

Text: le donne che parlano di calcio non si possono sentire.
Assumption: le donne non capiscono niente di calcio, è una cosa da maschi.
Answer: SÌ.

Text: Dopo aver visto (per sbaglio) 5 minuti di Uomini e Donne ho deciso che odierò le donne.
Assumption: Le donne che vanno alla trasmissione Uomini e Donne sono stupide.
Answer: SÌ.

Text: {text}
Assumption:
Answer:
\end{prompt}

\subsection{Italian Wikipedia EN}
\label{prompt-it-en-Wikipedie}
\begin{prompt}
You are a classifier for implicit misogyny.
What are the implied assumptions of the text {text}? 
Based on the implied assumptions and the context provided, is the text misogynous? Answer only with YES or NO. Return ONLY valid JSON. 

Text: non dovremmo abbassare i nostri standard solo per assumere più donne.
Context: Assunzione agevolata di donne disoccupate o in specifici settori: esonero dai contributi INPS per due anni sui contratti a tempo indeterminato stipulati dal 1° settembre 2024.
Assumption: le donne sono meno qualificate degli uomini.
Answer: SÌ.

Text: le donne che parlano di calcio non si possono sentire.
Context: Per la prima volta una donna è stata scelta come prima voce telecronista per i Mondiali 2026. 
Assumption: le donne non capiscono niente di calcio, è una cosa da maschi.
Answer: SÌ.

Text: Dopo aver visto (per sbaglio) 5 minuti di Uomini e Donne ho deciso che odierò le donne.
Context: "Ora basta, siete ridicoli" Uomini e Donne, durissimo sfogo dell'ex tronista Ida Platano: il motivo? Davvero clamoroso
Assumption: Le donne che vanno alla trasmissione Uomini e Donne sono stupide.
Answer: SÌ.

Testo: {text}
Contesto: {contesto}
Conoscenza: {retrieved_text}
Presupposto:

\end{prompt}

\subsection{Italian Slang}
\label{prompt-it-en-slang}
\begin{prompt}
You are a classifier for implicit misogyny.

Tasks:
1. You are given a text and a set of retrieved entries from Wiktionary containing English online slang definitions.
2. Identify the implied assumption in the text.
3. Based on the implied assumption, classify the text as misogynous or not. Answer only with YES or NO.

Return ONLY valid JSON.

Example:

Text: "Avrà mostrato le menne per ottenere la posizione"
Context: "Christina Koch, prima donna verso la Luna: chi è e cosa farà su Artemis II"
Slang: "menne"
Definition: "(dialectal slang) Women's breast"
Implied assumption: "Christina Koch ha ottenuto la posizione sfruttando il proprio corpo, piuttosto che grazie alle sue competenze e qualifiche professionali."
Answer: YES

Now process the following:

Text: {text}
Context: {Contesto}
Wiktionary entries:
{retrieved_text}
\end{prompt}

\subsection{Italian Literature EN}
\label{prompt-it-en-literature}
\begin{prompt}
You are a classifier for implicit misogyny.

Tasks:
1. You are given a text and a set of retrieved entries from gender studies books.
2. Identify the implied assumption in the text.
3. Based on the implied assumption and the literature, if relevant, classify the text as misogynous or not. Answer only with YES or NO.

Return ONLY valid JSON.

Examples:

Context: La festa in piscina finisce in stupro, notte horror per due 20enni. «Mi sono svegliata con un ragazzo sopra di me»
Text: Sinceramente non mi fa pena alla fine della storia sempre alcol e droga (non in questo caso) questi giovani devono trovare altre strade per divertirsi.
Literature: Codes applied included the presence or absence of physical injury, perpetrator's use of a weapon or force, the circumstances of the alleged assault (such as location, presence of drugs or alcohol) and the accuser's relationship to the alleged perpetrator (stranger, acquaintance, friend, boyfriend).
Implied assumption: Le donne che vengono stuprate vengono colpevolizzate se usano alcol o droghe
Answer: YES

Context: Femminicidi, un corteo e uno sciopero alla islandese. Ecco perché i social si stanno tingendo di rosso.
Text: Brutta storia essere aride e sterili femministe.
Literature: feminists that refuse to accept unjust societal conditions that include violence, harassment, silencing and other daily tactics of marginalization are depicted by different publics, from mainstream media to anti-feminist counterpublics, as disruptive and full of rage.
Implied assumption: Le femministe sono considerate isteriche e rabbiose.
Answer: YES

Now process the following:

Text: {text}
Context.{Contesto}
Literature retrieved: {retrieved_text}
\end{prompt}

\section{Retrieval Design Choices}
\label{app:similarity-thresholds}

\paragraph{Threshold selection.}
We selected the similarity threshold through a manual qualitative analysis of $200$ instances, examining evidence retained by each knowledge source at different threshold values. For each source we noticed that lower values with respect to the selected one retained a substantial amount of irrelevant and noisy information while the higher threshold with respect to the selected one removed evidence that was useful for the input. We therefore selected the intermediate values ($0.3$ for Wikipedia, $0.3$ for slang and $0.6$ for feminist-literature retrieval) as the best trade-off between retrieval relevance and information coverage.

\paragraph{Final sentence-level compression.}
For the second compression, we compared thresholds of 0.2,0.3 and 0.4. A threshold of 0.2 retained excessive noise, whereas a threshold of 0.4 removed too much potentially relevant information, therefore, we selected 0.3.

\paragraph{Compression vs. no compression}
To assess whether the second compression stage (Section~\ref{par:second-compression}) is useful, we compare recall on \textsc{ImplicIT-Mis} (Wikipedia, IT, few-shot) with and without this stage obtaining 0.881 without compression vs. 0.883 with compression, therefore compression slightly improves performance, and we retain it as it reduces the amount of noise in the retrieved evidence passed to the model.

\section{Results Across Model Sizes}
\label{app:model-sizes}

We additionally evaluate LLaMA-3.2-1B-Instruct and LLaMA-3.2-3B-Instruct to assess whether the retrieval effects observed with the main LLaMA-3.1-8B-Instruct change across different sizes of the same family of models on a selected subset of configurations (Table \ref{tab:smaller_models}). Results show that the effect of retrieval varies with model size. In some configurations, the 1B model benefits more from RAG than the 8B model. For instance, in the Italian few-shot baseline setting with Wikipedia retrieval and no second compression, recall increases from 0.754 without RAG to 0.949 with RAG. This may indicate that retrieved knowledge is more useful when the model has less parametric knowledge, although the pattern is not consistent across all settings therefore it would require further testing. 
\begin{table}[h]
    \centering
    \footnotesize
    \setlength{\tabcolsep}{3.5 pt}
    \begin{tabular}{llllrr}
    \toprule
    \textbf{Model} & \textbf{Data} & \textbf{KB} & \textbf{Setting} & \textbf{RAG} & \textbf{No-RAG}\\
    \midrule
    1B & IT & Wikipedia & few-shot & 0.949 & 0.754 \\
    1B & EN & Wikipedia & zero-shot & 0.887 & 0.882 \\
    3B & IT & Wikipedia & few-shot & 0.893 & 0.868 \\
    3B & EN & Wikipedia & zero-shot & 0.964 & 0.965 \\
    \midrule
    1B & EN & Slang & few-shot & 0.946 & 0.944 \\
    3B & IT & Slang & few-shot & 0.712 & 0.666 \\
    \bottomrule

    \end{tabular}
    \caption{Results for smaller models.}
    \label{tab:smaller_models}
\end{table}

\section{Human Evaluation Guidelines}
\label{app:guidelines}

\paragraph{Evaluation Goal} For each item, you are given four AI generated explanations. Compare them with respect to the following dimensions: \begin{itemize} \item \textbf{Faithfulness / groundedness} Whether the explanation is supported by the input text and, in the RAG conditions, by the retrieved context. \item \textbf{Human preference} Which explanation a human annotator would prefer, considering correctness, clarity and usefulness. \end{itemize} \subsection{1. Faithfulness / Groundedness} \paragraph{Definition} Faithfulness/groundedness measures whether the generated explanation is justified by the available evidence. You can see an example and the generated explanation based on the score. \paragraph{Example} \textbf{Sentence:} \textit{my wife gets angry when i call her pet names apparently sandwich making machine is sexist. . .} \paragraph{2 --- Fully grounded} The explanation accurately identifies the implicit misogyny and is clearly supported by the input and/or retrieved context. \textbf{Generated:} \textit{The sentence is sarcastic and “sandwich making machine” relies on a sexist stereotype that reduces women to domestic labor.} \paragraph{1 --- Partially grounded} The explanation captures part of the intended meaning but is incomplete, vague, or only weakly connected to the evidence. \textbf{Example} \textbf{Sentence:} \textit{my wife gets angry when i call her pet names apparently sandwich making machine is sexist. . .} \textbf{Generated:} \textit{The text is sexist because it refers to the speaker’s wife in a disrespectful way.} \paragraph{0 --- Not grounded / hallucinated} The explanation is unsupported, contradicts the text, or invents content. \textbf{Example} \textbf{Sentence:} \textit{my wife gets angry when i call her pet names apparently sandwich making machine is sexist. . .} \textbf{Explanation:} \textit{The text mentions a personal interaction between the speaker and their wife, and does not make a generalization about women.} 

\subsection{2. Human Preference} 
\paragraph{Definition} 
Human preference measures which explanation annotators find better overall. They should consider: \begin{itemize} \item correctness; \item clarity; \item usefulness for understanding implicit misogyny; \item avoidance of hallucinated or overgeneralized claims. \end{itemize} 
\paragraph{Preference Labels} 
Use one of the following labels: 

\begin{center}
\footnotesize
\begin{tabular}{lp{0.7\linewidth}}
\toprule
\textbf{Label} & \textbf{Meaning} \\
\midrule
Prefer \textless l\textgreater & Explanation \textless l\textgreater\ is clearly better than the others \\
Tie & Some or all explanations are similarly good or similarly flawed \\
None & All explanations are wrong \\
\bottomrule
\end{tabular}
\end{center}

\subsection{Inter-Annotator Agreement}
Table \ref{tab:faithfulness_agreement} shows the inter-annotator agreement for faithfulness.
\begin{table}[h]
\footnotesize
\centering
\begin{tabular}{|l|c|c|}
\hline
\textbf{Subset} & \textbf{Italian} & \textbf{English} \\
 & \textbf{Cohen's $\kappa$} & \textbf{Cohen's $\kappa$} \\
\hline
General    & 0.505 & 0.454 \\
Wiki       & 0.710 & 0.452 \\
Slang      & 0.680 & 0.504 \\
Literature & 0.650 & 0.421 \\
Baseline   & 0.700 & 0.437 \\
Human Preference & 0.286 & 0.333 \\
\hline
\end{tabular}
\caption{Inter-annotator agreement on Faithfulness: Cohen's $\kappa$ across subsets and languages.}
\label{tab:faithfulness_agreement}
\end{table}

\subsection{Annotators}
Human evaluations have been carried out by two authors of this paper. Annotator 1 is a Postdoctoral researcher identifying as a woman, expert in online misogyny. She is also a feminist activist. Her background is in humanities and NLP.
Annotator 2 is a Master's student in Data Science and Business Analytics identifying as a woman. They first annotated independently. They then discussed the disagreement cases and clarified the interpretation of the annotation guidelines before independently re-annotating the disputed cases.

\end{document}